%% file: camera_ready.tex
\documentclass[10pt,twocolumn,letterpaper]{article}

\usepackage{cvpr}              

\usepackage{graphicx}
\usepackage{amsmath}
\usepackage{amssymb}
\usepackage{booktabs}
\usepackage{bm, amssymb, amsfonts}
\usepackage{array}

\usepackage[accsupp]{axessibility} 

\usepackage[pagebackref,breaklinks,colorlinks]{hyperref}

\usepackage{dsfont}
\usepackage{multirow}
\usepackage{stmaryrd}
\usepackage{booktabs}
\usepackage{nicefrac}
\usepackage{relsize}
\usepackage{mathabx}
\usepackage{subcaption}
\usepackage{xspace}
\usepackage{multirow}
\usepackage{array}
\usepackage[dvipsnames]{xcolor}
\usepackage{mathtools}
\usepackage{tikz}
\usepackage{array}
\newcolumntype{H}{>{\setbox0=\hbox\bgroup}c<{\egroup}@{}}
\usepackage{xspace}
\usepackage{pifont}
\usepackage{tabularx}
\makeatletter
\DeclareRobustCommand\onedot{\futurelet\@let@token\@onedot}
\def\@onedot{\ifx\@let@token.\else.\null\fi\xspace}
\def\eg{\emph{e.g}\onedot} 
\def\ie{\emph{i.e}\onedot}

\def\wrt{\emph{w.r.t}\onedot}

\makeatother

\usepackage[capitalize]{cleveref}
\crefname{section}{Sec.}{Secs.}
\Crefname{section}{Section}{Sections}
\Crefname{table}{Table}{Tables}
\crefname{table}{Tab.}{Tabs.}

\usepackage{mathtools}
\usepackage{leftindex}
\usepackage{float}
\usepackage{svg}
\usepackage{stackengine}
\def\delequal{\mathrel{\ensurestackMath{\stackon[1pt]{=}{\scriptscriptstyle\Delta}}}}
\usepackage{bbm}

\usepackage[normalem]{ulem}
\useunder{\uline}{\ul}{}

\def\cvprPaperID{13204} 
\def\confName{CVPR}
\def\confYear{2024}

\def\true{1}
\def\showcomments{1} 
\ifx\showcomments\true
  \newcommand{\comments}[1]{\textcolor{blue}{#1}}
\else
   \newcommand{\comments}[1]{}
\fi

\begin{document}
\input{macro.tex}


\title{MultiPhys: Multi-Person Physics-aware 3D Motion Estimation}

\author{%
 Nicolas Ugrinovic\thanks{Work done during internship at Stanford.}\,\,$^{1,2}$ \quad
 Boxiao Pan$^{2}$ \quad
 Georgios Pavlakos$^{3}$ \quad Despoina Paschalidou$^{2}$ \\
 Bokui Shen$^{2}$ \quad \quad Jordi Sanchez-Riera$^{1}$\quad
  Francesc Moreno-Noguer$^{1}$\quad Leonidas Guibas$^{2}$\\
  $^1$Institut de Robotica i Informatica Industrial, CSIC-UPC, Barcelona, Spain\\
  $^2$Stanford University \quad $^3$UT Austin\\
}

\input{figures/teaser}

\def\thefootnote{*}\footnotetext{Work done during internship at Stanford.}\def\thefootnote{\arabic{footnote}}
\input{01_abstract}

\input{02_intro}
\input{03_related_work}

\input{04_methodology}

\input{05_experiments}

\input{06_conclusion}

\section*{Acknowledgments}
Despoina Paschalidou is supported by the Swiss National Science Foundation
under grant number P500PT\_206946. Leonidas Guibas is supported by a Vannevar
Bush Faculty Fellowship. This work is partially supported by   projects SMARTGAZEII CPP2021-008760 and MoHuCo PID2020-120049RB-I00.

{\small
\bibliographystyle{ieee_fullname}
\bibliography{references}
}

\end{document}

%% file: macro.tex
\newcommand{\GA}[1]{{\color{violet}#1}}
\newcommand{\GPM}[1]{{\color{blue} GPM:#1}} 

\newcommand{\mA}{\mathcal{A}}
\newcommand{\mF}{\mathcal{F}}
\newcommand{\mI}{\mathcal{I}}
\newcommand{\mK}{\mathcal{K}}
\newcommand{\mP}{\mathcal{P}}
\newcommand{\mR}{\mathcal{R}}
\newcommand{\mU}{\mathcal{U}}
\newcommand{\mZ}{\mathcal{Z}}

\newcommand{\bA}{\mathbf{A}}
\newcommand{\bB}{\mathbf{B}}
\newcommand{\bC}{\mathbf{C}}
\newcommand{\bD}{\mathbf{D}}
\newcommand{\bF}{\mathbf{F}}
\newcommand{\btF}{\tilde{\mathbf{F}}}
\newcommand{\bG}{\mathbf{G}}
\newcommand{\bH}{\mathbf{H}}
\newcommand{\bI}{\mathbf{I}}
\newcommand{\bJ}{\mathbf{J}}
\newcommand{\bK}{\mathbf{K}}
\newcommand{\bL}{\mathbf{L}}
\newcommand{\bM}{\mathbf{M}}
\newcommand{\bN}{\mathbf{N}}
\newcommand{\bO}{\mathbf{O}}
\newcommand{\bP}{\mathbf{P}}
\newcommand{\btP}{\tilde{\mathbf{P}}}
\newcommand{\bQ}{\mathbf{Q}}
\newcommand{\bR}{\mathbf{R}}
\newcommand{\btR}{\tilde{\mathbf{R}}}
\newcommand{\btS}{\tilde{\mathbf{S}}}
\newcommand{\bS}{\mathbf{S}}
\newcommand{\bT}{\mathbf{T}}
\newcommand{\btT}{\tilde{\mathbf{T}}}
\newcommand{\bU}{\mathbf{U}}
\newcommand{\bV}{\mathbf{V}}
\newcommand{\btV}{\tilde{\mathbf{V}}}
\newcommand{\bW}{\mathbf{W}}
\newcommand{\bX}{\mathbf{X}}
\newcommand{\btX}{\tilde{\mathbf{X}}}
\newcommand{\bY}{\mathbf{Y}}
\newcommand{\btY}{\tilde{\mathbf{Y}}}
\newcommand{\bZ}{\mathbf{Z}}

\newcommand{\bzero}{\textbf{0}}
\newcommand{\bone}{\textbf{1}}
\newcommand{\ba}{\mathbf{a}}
\newcommand{\bb}{\mathbf{b}}
\newcommand{\bc}{\mathbf{c}}
\newcommand{\bd}{\mathbf{d}}
\newcommand{\be}{\mathbf{e}}
\newcommand{\bff}{\mathbf{f}}
\newcommand{\bg}{\mathbf{g}}
\newcommand{\bh}{\mathbf{h}}
\newcommand{\bi}{\mathbf{i}}
\newcommand{\bj}{\mathbf{j}}
\newcommand{\bl}{\mathbf{l}}
\newcommand{\bn}{\mathbf{n}}
\newcommand{\bo}{\mathbf{o}}
\newcommand{\bp}{\mathbf{p}}
\newcommand{\bq}{\mathbf{q}}
\newcommand{\br}{\mathbf{r}}
\newcommand{\bs}{\mathbf{s}}
\newcommand{\bts}{\tilde{\mathbf{s}}}
\newcommand{\bt}{\mathbf{t}}
\newcommand{\bu}{\mathbf{u}}
\newcommand{\btu}{\tilde{\mathbf{u}}}
\newcommand{\bv}{\mathbf{v}}
\newcommand{\btv}{\tilde{\mathbf{v}}}
\newcommand{\bbv}{\bar{\mathbf{v}}}
\newcommand{\bw}{\mathbf{w}}
\newcommand{\bx}{\mathbf{x}}
\newcommand{\btx}{\tilde{\mathbf{x}}}
\newcommand{\by}{\mathbf{y}}
\newcommand{\bty}{\tilde{\mathbf{y}}}
\newcommand{\bz}{\mathbf{z}}
\newcommand{\btz}{\tilde{\mathbf{z}}}
\newcommand{\bhz}{\hat{\mathbf{z}}}

\renewcommand{\vec}[1]{\boldsymbol{#1}}
\newcommand{\mat}[1]{\mathbf{#1}}
\newcommand{\set}[1]{\mathcal{#1}}

\newcommand{\real}[0]{\mathbb{R}}
\newcommand{\tb}[0]{\textbf}
\newcommand{\ti}[0]{\textit}
\newcommand{\et}[0]{\ti{et al.}}

\newcommand{\imageset}[0]{\set{I}}
\newcommand{\image}[0]{\mat{I}}

\newcommand{\poseset}[0]{\set{P}}
\newcommand{\transset}[0]{\set{T}}
\newcommand{\jointset}[0]{\set{J}}
\newcommand{\garmparamset}[0]{\set{G}}

\newcommand{\template}[0]{\mat{T}}
\newcommand{\garment}[0]{\mat{G}}
\newcommand{\blendweight}[0]{w}
\newcommand{\blendweights}[0]{\mat{W}}

\newcommand{\normal}[0]{{\mathbf{n}}}

\newcommand{\depth}[0]{\widehat{\vec{d}}}
\newcommand{\ankl}[0]{\mathbf{x}_{l}}
\newcommand{\ankr}[0]{\mathbf{x}_{r}}
\newcommand{\lplane}[0]{L_{p}}
\newcommand{\lamdp}[0]{\lambda_{p}}

\newcommand{\joints}[0]{\mat{J}}
\newcommand{\jointsTwoD}[0]{\tilde{\mat{J}}}

\newcommand{\rot}[0]{\vec{R}}

\newcommand{\scale}[0]{\mat{s}}

\newcommand{\nplane}[0]{\bf{\widehat{{n}}} }

\newcommand{\garmparam}[0]{\vec{z}}
\newcommand{\offsets}[0]{\mathbf{D}}

\newcommand{\cut}{\mat{z}_\mathrm{cut}}
\newcommand{\style}{\mat{z}_\mathrm{style}}
\newcommand{\zpose}{\mat{z}_{\pose}}

\newcommand{\smpl}[0]{M}
\newcommand{\posefun}[0]{T}
\newcommand{\blendfun}[0]{W}
\newcommand{\offsetfun}[0]{B}
\newcommand{\offsetsfun}[0]{D}
\newcommand{\jointfun}[0]{J}
\newcommand{\garmfun}[0]{G}

\newcommand{\metricdist}{pairwise normalized distances between persons}

\newcommand{\numberOfMetrics}{three} 

\newcommand{\Modelname}{Keep Your Feet on the Ground} 

\newcommand{\lossRep}[0]{L}
\newcommand{\multilossRep}[0]{\hat{L}}
\newcommand{\lambdaRep}[0]{\lambda_{2D}}

\newcommand{\heightDistMetric}[0]{h_{err}}

\newcommand{\posevect}[0]{\vec{p}}
\newcommand{\posevectRef}[0]{\vec{q}}
\newcommand{\jointsDims}[0]{\mathbb{R}^{J\times3}}
\newcommand{\diffVect}[0]{\vec{\delta}}

\newcommand{\featureVect}[0]{\vec{f}}
\newcommand{\featureNet}[0]{\mathcal{G}}
\newcommand{\ourNet}[0]{\Phi}
\newcommand{\rfineNet}[0]{\mathcal{R}}

\newcommand{\reals}[0]{\mathbb{R}}

\newcommand{\xmark}{\ding{55}}%

\newcommand{\kinpose}{\widetilde{\mathbf{q}}^{i}_{t}}
\newcommand{\kinposeW}{{}^{w}\kinpose}
\newcommand{\kinposeS}{{}^{s}\kinpose}

\newcommand{\simpose}{\mathbf{q}^{i}_{t}}
\newcommand{\simposeW}{{}^{w}\simpose}
\newcommand{\simposeS}{{}^{s}\simpose}

\newcommand{\ori}[1][\Phi]{{#1}^{i}_{t}}
\newcommand{\oriW}{{}^{w}\ori}
\newcommand{\oriS}{{}^{s}\ori}
\newcommand{\oriKin}{\ori[\widetilde{\Phi}]}

\newcommand{\body}[1][\Theta]{{#1}^{i}_{t}}
\newcommand{\bodyW}{{}^{w}\body}
\newcommand{\bodyKin}{\body[\widetilde{\Theta}]}

\newcommand{\shape}{\beta^{i}}

\newcommand{\trans}[1][\Gamma]{{#1}^{i}_{t}}
\newcommand{\transW}{{}^{w}\trans}
\newcommand{\transS}{{}^{s}\trans}
\newcommand{\transKin}{\trans[\widetilde{\Gamma}]}

\newcommand{\scit}[2][t]{{#2}^{i}_{#1}}
\newcommand{\plusbinomial}[3][2]{(#2 + #3)^#1}

\newcommand{\simOri}{\scit{R}}
\newcommand{\simTr}{\scit{T}}
\newcommand{\simBody}{\scit{\Omega}}

\newcommand{\qpos}[1]{\scit[#1]{\mathbf{q}}}
\newcommand{\qposKin}[1]{\scit[#1]{\widetilde{\mathbf{q}}}}

\newcommand{\action}{\scit{\mathbf{a}}}
\newcommand{\curS}{\scit{\mathbf{s}}}
\newcommand{\tgPose}{\scit[t+1]{\widetilde{\mathbf{q}}}}

\newcommand{\loopN}{\textit{loop-N }}

\newcommand{\scK}[2]{{}^{#2}{#1}}

\newcommand{\nicolas}[1]{{\textcolor{blue}{#1}}}
\newcommand{\boxiao}[1]{{\textcolor{red}{\textbf{Boxiao}: #1}}}
\newcommand{\george}[1]{{\textcolor{green}{\textbf{George}: #1}}}
\newcommand{\despi}[1]{{\textcolor{orange}{\textbf{Despi}: #1}}}
\newcommand{\nico}[1]{{\textcolor{purple}{\textbf{Nico}: #1}}}

\newcommand{\francesc}[1]{{\color{magenta}#1}} 
\newcommand{\francescrmk}[1]{{\color{magenta} {[\bf fmn: #1]}}}

\newcommand{\doA}{$\downarrow$}
\newcommand{\physics}{physics-based }
\newcommand{\ours}{MultiPhys}
\newcommand{\nloop}{N_{l} }
\newcommand{\pa}{PA-MPJPE  }
\newcommand{\wpe}{W-MPJPE  }

\newcommand{\projpage}{\href{http://www.iri.upc.edu/people/nugrinovic/multiphys/}{project page}}%

%% file: figures/teaser.tex
\twocolumn[{%
\renewcommand\twocolumn[1][]{#1} %

\maketitle
\thispagestyle{empty}

\begin{center}
    \centering
    \vspace*{0.3cm}
    \includegraphics[width=1.0\linewidth]{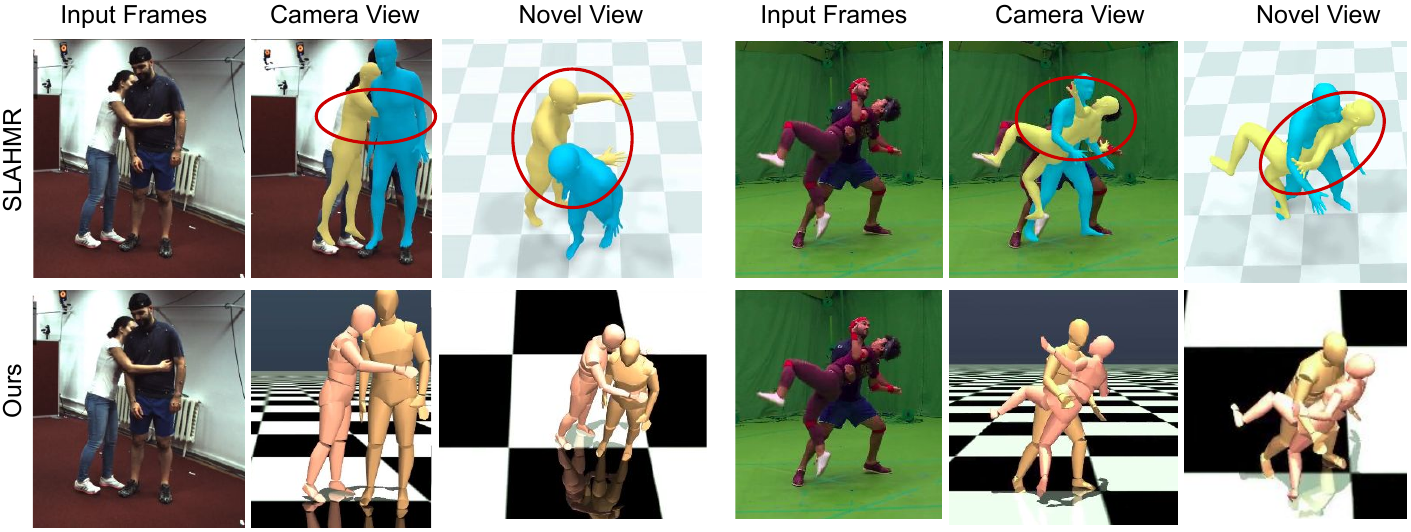} \\[-0.3cm]
    \vspace*{0.3cm}
    \captionof{figure}{\small{\textbf{MultiPhys enables recovering multi-person 3D motion in a physically-aware manner.}.  State-of-the-art methods (SLAHMR~\cite{ye2023slahmr}, top row) for multi-person motion recovery mostly rely on kinematic approaches, which typically ignore physical constraints, such as body penetration. 
    Note that while individual poses are kinematically coherent, their spatial placement is suboptimal, resulting in significant penetration errors. MultiPhys (bottom row) incorporates physics constraints into the reconstruction process, yielding more physically plausible results.}
    }\label{fig:teaser}
\end{center}
}]
\vspace*{-0.0cm}

%% file: 01_abstract.tex

\begin{abstract}
\vspace{-3mm}
We introduce \ours, a method designed for recovering multi-person motion from monocular videos. Our focus lies in capturing coherent spatial placement between pairs of individuals across varying degrees of engagement. MultiPhys, being physically aware, exhibits robustness to jittering and occlusions, and effectively eliminates penetration issues between the two individuals. We devise a pipeline in which the motion estimated by a kinematic-based method is fed into a physics simulator in an autoregressive manner. We introduce distinct components that enable our model to harness the simulator's properties without compromising the accuracy of the kinematic estimates. This results in final motion estimates that are both kinematically coherent and physically compliant.
Extensive evaluations on three challenging datasets characterized by substantial inter-person interaction show that our method significantly reduces errors associated with penetration and foot skating, while performing competitively with the state-of-the-art on motion accuracy and smoothness.
Results and code can be found in our \projpage. 

\end{abstract}

%% file: 02_intro.tex

\vspace*{-0.2cm}

\section{Introduction}

\label{sec:intro}

In recent years, significant advancements have been made in recovering human motion from monocular RGB videos~\cite{yuan2021simpoe,Luo2021kinpoly,Luo2022EmbodiedSH,yuan2022glamr,ye2023slahmr,HMR22023}. While most of these works focus on videos of a single person, estimating motions for multiple people, especially those interacting, becomes significantly more challenging. This challenge primarily arises due to severe occlusion during close interactions, leading to multiple body parts being invisible for extended periods.~\cref{fig:teaser} illustrates an example of this, causing previous state-of-the-art~\cite{ye2023slahmr} to produce motions with heavy inter-person penetrations. Additionally, previous methods also suffer from problems such as ground penetration and foot skating~\cite{yuan2022glamr,ye2023slahmr,HMR22023}. We argue that this is due to a lack of physics modeling.

In contrast, prior works in single-person motion estimation have explored incorporating physics into the process~\cite{Luo2021kinpoly,yuan2021simpoe,Luo2022EmbodiedSH}. These methods typically employ a physics simulator and train a motion policy to generate motion that complies with physical constraints while imitating input observations. However, extending such methods to the scenario of multiple people presents significant challenges. Due to severe occlusion, detected 2D keypoints become less reliable and methods that rely on them are prone to fail~\cite{yuan2021simpoe,Luo2022EmbodiedSH}. This raises the question of what input representation should be used. 

To this end, we propose a framework, dubbed \ours, that employs a physics simulation engine~\cite{todorov2012mujoco} to recover motion for multiple interacting people in a physics-plausible manner. Instead of relying on detected 2D keypoints in a feedforward model~\cite{yuan2021simpoe,Luo2022EmbodiedSH}, we initialize our approach with the output from SLAHMR~\cite{ye2023slahmr}, which performs global optimization on the entire sequence. However, since SLAHMR is physics-agnostic, its outputs may be noisy, particularly concerning the spatial placement of the bodies. This also results in inter-person penetrations. We devise a pipeline in which these preliminary body poses are fed into the physics simulator in an autoregressive fashion, aiming to obtain physically compliant motion estimates. 

We observe, however, that naively feeding these poses makes it difficult for the policy to generate the control signal to drive the simulation, which results in motion degradation. This happens especially when dealing with highly dynamic motions, as they present larger pose displacements between consecutive frames. As a result, the policy struggles to "catch up" with the reference signal.
To counter this, we design an iterative refinement procedure and observe that the input poses are better matched while remaining physically compliant. 

We evaluate \ours~on three challenging datasets. Our method performs competitively with previous state-of-the-arts on pose accuracy and smoothness, while significantly reducing errors on inter-body and ground penetration as well as foot skating.
 
Specifically, compared to SLAHMR~\cite{ye2023slahmr}, we improve the (inter-body) penetration score by more than 7 times across all datasets, and the ground penetration score by 30, 5, and 1.35 times on the three datasets, respectively. We also reduce skating by 35\% and 137\% on the Hi4D and ExPI datasets, respectively. 

In summary, our contributions are (1) A physics simulator-based framework for multi-person 3D motion estimation in a physically plausible manner. To the best of our knowledge, our method is the first that incorporates a physics simulator for multi-person 3D motion estimation; (2) Extensive evaluation showing that our method achieves significantly better results on physics-related metrics, while performing on par with prior works w.r.t. pose accuracy and smoothness.

%% file: 03_related_work.tex
\section{Related work}
\label{sec:related_work}

\noindent \textbf{Human mesh recovery from videos.}
Methods that reconstruct human mesh from videos~\cite{hmmr,VIBECVPR2020} build upon earlier works that focus on single images~\cite{kanazawa_hmr}, incorporating temporal coherence to enhance motion reconstruction. Consequently, these methods are able to recover smooth and plausible human motion. While these and other regression methods that follow the same line of work~\cite{choi2021beyond,luo2020_motion,pavlakos2022multishot} make valuable contributions, they often lack the ability to recover global trajectory -- a crucial aspect for a comprehensive understanding of human motion. Simply extending these methods to videos with multiple people leads to spatially incoherent distribution of human meshes. To mitigate this issue, recent approaches focus on estimating global trajectories from per-frame local human poses~\cite{yuan2022glamr,li2022dnd,MovingCam2021}. Others take a step further by incorporating motion cues and additional constraints to more faithfully track the global trajectory ~\cite{TRACE,ye2023slahmr,kocabas2024pace}. TRACE~\cite{TRACE} uses optical flow cues to track human motions. SLAHMR~\cite{ye2023slahmr} and PACE~\cite{kocabas2024pace} employ SLAM to compute dynamic camera parameters, refining them in an optimization stage. 
These methods also include constraints such as motion priors and contact with an estimated ground plane. Despite their impressive results, they often overlook fundamental physical constraints governing human motion in the real world, such as gravity and collisions with other individuals. To address these limitations, we propose to enforce these constraints explicitly by leveraging a full-featured physics engine.

\noindent \textbf{Physics-based approaches.}
Trying to cope with the limitation that most human mesh recovery methods do not include physic constraints, some works explicitly incorporate physics notions. HuMoR~\cite{humor}, while not a physics simulator-based approach, models the human motion dynamics using a probabilistic generative prior that is later used in an optimization framework to recover motion. In the optimization stage, some losses for foot skating and velocity are applied to force physical compliance. However, once again, these are applied as soft constraints. Others~\cite{gartner2022trajectory,shimadaPhysCap} take a step further and incorporate a physics simulator into their pipeline to reconstruct single-person motion. However, since most full-featured physics simulators are not differentiable, alternative optimization methods must be explored. For instance,~\cite{gartner2022trajectory} utilizes an evolutionary algorithm for refining poses, which can be challenging to optimize and result in extended inference times. To address this,~\cite{gartner2022diffphy,neuralPhysCap} use differentiable physics. While this is a promising approach, differentiable simulators either require specific formulations for each problem or are often simplistic, lacking the features of non-differentiable simulators. 
Finally, human motion capture methods from sparse IMUs turn to utilizing physical simulators~\cite{winkler2022questsim,lee2023questenvsim} or apply soft physical constraints~\cite{zheng2023realistic,jiang2022avatarposer}. 
All these approaches focus on a single person, whereas our method handles multiple people.

\noindent \textbf{RL approaches.}
Works that use Reinforcement Learning (RL) to reconstruct human motion include~\cite{yuan2021simpoe,yuan2020residual,Luo2022EmbodiedSH,Luo2021kinpoly,2018-TOG-SFV}. Typically, these approaches involve learning a control policy in a simulation environment to govern a humanoid agent. Subsequently, another policy is often learned to generate reference motion, operating in conjunction with the first policy. These methods often leverage fully-featured simulators and utilize more realistic humanoid models compared to other works. Our work builds upon~\cite{Luo2022EmbodiedSH}, originally designed for a single person.

\noindent \textbf{Multi-person human mesh recovery.}
Several approaches address multi-person mesh recovery, each targeting specific challenges within this problem, such as occlusion~\cite{Khirodkar_2022_CVPR,ROMP}, depth ambiguity~\cite{ugrinovic2021depthsize}, or accurate spatial placement~\cite{BEV,jiang2020_multiperson}. A limited number of methods have concentrated on modeling contacts and addressing penetration issues~\cite{jiang2020_multiperson,Fieraru_2020_CVPR,muller2023buddi,REMIPS}. While these existing works focus on recovering human mesh solely from static images, our proposed method takes a step further by estimating multiple humans from videos while focusing on addressing the ground and inter-person penetration, as well as foot skating issues. There exist other methods that use multi-view cameras for capturing multi-person motion~\cite{multiHumanData}, which is out of the scope of this work.

%% file: 04_methodology.tex
\section{Method}
\label{sec:meth}

\input{figures/overview}

In the following, we describe the steps we take to correct an initial kinematic motion estimate to be physically plausible. We first introduce the necessary background on SLAHMR in~\cref{subsec:slahmr}. We then formalize our problem in~\cref{subsec:problem}. Next, we present an overview of our pipeline in~\cref{subsec:overview}. Finally in~\cref{sec:meth:simu}, we describe in detail the core part of our method, which is the physics-aware correction. 

\subsection{Background on SLAHMR~\cite{ye2023slahmr}}
\label{subsec:slahmr}
SLAHMR~\cite{ye2023slahmr} offers the state-of-the-art solution for multi-person motion estimation in a global coordinate frame. SLAHMR proposes a multi-stage optimization-based pipeline, where it optimizes a number of objectives as soft constraints. These include reprojection, motion smoothness, foot skating, ground contacts, pose and motion priors. This method is relatively robust to occlusion, thanks to the pose~\cite{smplx} and motion~\cite{humor} priors it leverages. However, since it does not explicitly model physics, the estimated motion often manifests severe inter-person and ground penetration, as well as foot skating. In this work, we take the motion estimate from SLAHMR as the initialization and correct it to be physically plausible via our proposed method.

\subsection{Problem Formulation}
\label{subsec:problem}
Given a monocular video $I_{1:T}$ consisting of $T$ frames that have $N$ interacting people, our goal is to recover the motion in world coordinates in a \textit{physically plausible} manner, denoted as $\mathbf{Q}^i = \{\mathbf{q}^i_0, \mathbf{q}^i_1, \ldots, \mathbf{q}^i_T\}$, for all people $i = 1 \ldots N$. Each pose is parameterized following the SMPL-H~\cite{smplh2022} body model, which consists of the global orientation $\ori \in \real^3$, body pose $\body \in \real^{22 \times 3}$, body shape $\shape \in \real^{16}$, and root translation $\trans \in \real^3$. That is: 
\begin{equation}
\simpose = \{\ori,\body,\shape,\trans\}
\end{equation}
Throughout our experiments, $N = 2$, but note that our method does not have a fundamental limit on $N$ and can in theory work with an arbitrary number of interacting people.

\subsection{Method Overview}
\label{subsec:overview}
An overview of our method is presented in~\cref{fig:overview}.
We use a set of poses estimated by SLAHMR~\cite{ye2023slahmr}, denoted by $\kinpose$, as the initial estimates, which are later corrected to be physically plausible using a physics simulator. We denote these corrected poses as $\simpose$. Here each pose is for the $i$-th person at timestep $t$. 

The motion estimated from SLAHMR $\kinpose$ is also represented with the SMPL-H model:
\begin{equation}
\kinpose = \{\oriKin,\bodyKin,\shape,\transKin\}
\end{equation}
Note that the body shape $\beta$ is the same for all timesteps and we directly keep it from the SLAHMR estimates.

To enforce the physical constraints missing from SLAHMR, we leverage a full-featured physics simulator (Mujoco~\cite{todorov2012mujoco}). Inside the simulation, we represent each person as a humanoid agent consisting of different body parts and joints with actuators over those joints. We create one humanoid for each person. The creation process follows SimPoE~\cite{yuan2021simpoe}.

To prevent the simulated character from losing track of the input motion, we re-purpose the Universal Humanoid Controller (UHC)~\cite{Luo2022EmbodiedSH}, whose goal was to imitate a set of target poses while producing signals to drive the simulated character. In the original paper, the target poses input to the UHC are parameterized by the proposed Multi-step Projection Gradients (MPG), which link the 2D observations to 3D simulated body poses via gradients of the 2D reprojection loss. In our case, we re-purpose the UHC to take the SLAHMR-estimated poses $\kinpose$ as input. Note that the UHC also receives as input the body shape, thus different bodies can be controlled. This is especially important when working with multiple interacting people.

\input{figures/method}

\subsection{Physics-Aware Correction}
\label{sec:meth:simu}

A detailed diagram of our physics-aware correction is presented in~\cref{fig:simu}.
For this stage, our formulation is the following. Inside the simulation, we define the 3D human pose as $\simpose$.
At each timestep, we have access to each agent's state $\scit{\mathbf{s}} \delequal (\scit{\mathbf{q}}, \scit{\dot{\mathbf{q}}})$ which is the combination of the 3D pose and velocity. 

In order to drive the humanoid agents inside the simulation to mimic the kinematic poses and thus correct them to be physically plausible, we use an imitation policy similar to~\cite{Luo2022EmbodiedSH}. 
This policy is modeled as a Markov Decision Process (MDP) following the standard formulation in physics-based character control. This process is defined as a tuple $\mathcal{M}=\langle \mathcal{S},\mathcal{A},\mathcal{T},\mathcal{R},\gamma \rangle$ of states, actions, transition dynamics, reward function, and discount factor. The state $s \in \mathcal{S}$, reward $r \in \mathcal{R}$, and the transition dynamics $\mathcal{T}$ are determined by the physics simulator, while the action $a \in \mathcal{A}$ is given by the control policy $\pi(\action|\curS,\qposKin{t+1},\shape)$. We employ Proximal Policy Optimization~\cite{schulman2017proximal} (PPO) to find the the optimal control policy $\pi$ that maximizes the expected discounted reward $\mathbbm{E}[\sum^{T}_{t=1}\gamma^{t-1}r_t]$.

The connection between the kinematic estimates and the simulation is through the policy $\pi$, as shown in~\cref{fig:simu}. We drive the simulation with the kinematic poses $\kinpose$ by inputting them into the policy. In this way, the agent in the simulation will follow these poses as the reference.
We use the same policy for each agent in the scene. Each agent is controlled independently, mimicking what happens in reality (\ie each person can move independently in the world), but they all reside in the same simulation. By having different agents sharing the simulation simultaneously, we can directly impose physical restrictions between them, \eg they cannot penetrate each other. This results in physically compliant estimates. This formulation also has practical implications that we do not need to train the policy on multi-person datasets. Instead, it is trained only on the large-scale single-person MoCap dataset, AMASS~\cite{AMASS:ICCV:2019}.

For motion sequences where the persons are moving a lot and especially when they undergo extreme poses (\eg in the ExPI~\cite{guo2021multi} dataset), we observe that it is more difficult for the policy to imitate the reference poses. This happens because the policy was trained with a specific set of actions and poses. For any action that has different dynamics and overall different distribution from the training data, the policy is not able to completely match the target pose. Given that the policy's formulation follows the form of an auto-regressive system, this error tends to accumulate quickly, leading to noticeable final errors. To remediate this, we devise an iterative strategy for the agent to slowly get closer to the reference pose. Throughout the paper, we dub this strategy \loopN. As stated before, the policy samples an action $\action$ for each agent $i$ at timestep $t$ given the current simulation state of each agent ($\curS$) and reference pose ($\tgPose$). In normal operation, the updated 3D pose taken from the simulation output once the action is applied is defined by:
\begin{equation}
\label{eq:update}
     \qpos{t+1} = \mathcal{T}(\qpos{t},\action).
\end{equation}

To help $\qpos{t+1}$ match the reference pose, we iteratively update the simulation state for $\nloop$ iterations while keeping the reference pose fixed until it gets close enough to the it. Let $k$ be the current iteration while keeping the reference pose fixed, where $k=\{1,2,...,\nloop\}$. For every iteration, we sample a new action $\pi(\scK{\action}{k}|\scK{\curS}{k},\qposKin{t+1},\shape)$ that will drive the current pose closer to the reference pose. For the updates inside "\loopN", we redefine Eq.~\ref{eq:update} in terms of $k$:
\begin{equation}
     \scK{\qpos{t}}{k+1} = \mathcal{T}(\scK{\qpos{t}}{k} ,\scK{\action}{k}).
\end{equation}

After $\nloop$ iterations, we keep only the last update, which should be closer to the reference pose. Specifically, we make $\qpos{t+1}=\scK{\qpos{t}}{\nloop}$. We repeat this process for every timestep in the sequence. Once we project all the sequences into a physically plausible space, it is trivial to convert these poses back to the SMPL representation.

%% file: figures/overview.tex
\begin{figure*}[]
\centering
\includegraphics[width=1\linewidth]{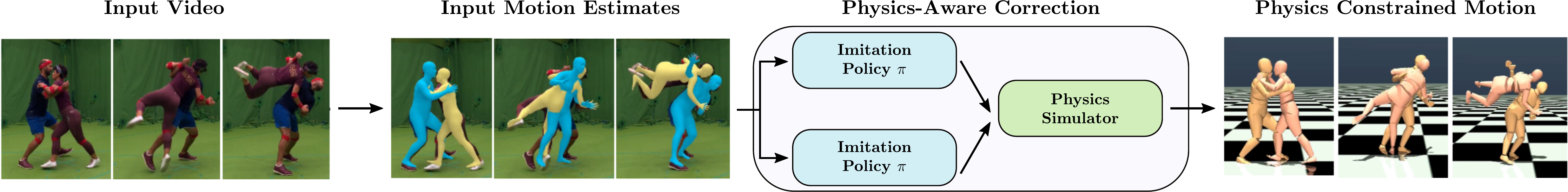} \\[-0.2cm]
\caption{
\textbf{\ours~Pipeline.} Given an input video with multiple people (left), we first obtain initial kinematic estimates of the camera poses and 3D human motion using SLAHMR~\cite{ye2023slahmr}. Using these initial motion estimates, our proposed framework corrects them and makes them physically plausible (right). 
\vspace*{-0.2cm}
}
\label{fig:overview}
\end{figure*}

%% file: figures/method.tex
\begin{figure}
\centering
\includegraphics[width=1\linewidth]{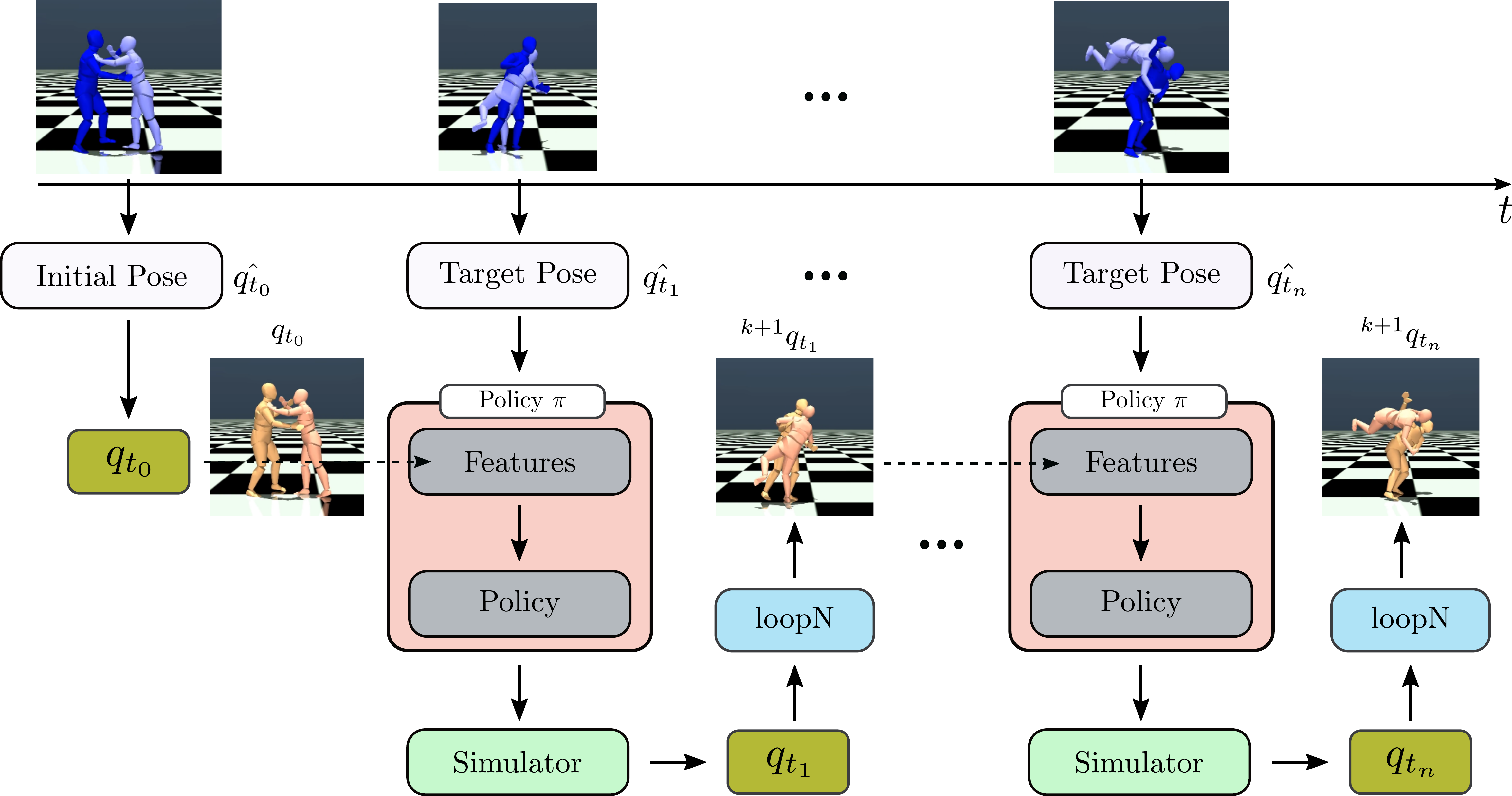} \\[-0.2cm]
\vspace*{0.2cm}
\caption{
\textbf{Physics-aware Correction Module}. We use the policy $\pi$ to control the humanoid agents with the initial kinematic poses. We simulate all agents simultaneously in order to apply physics-based constraints to the reconstructed motion. The policy computes features from both the current state of the simulation and the target pose to later generate the action signal $a$ that controls the agents. We place our \loopN component between target poses $\tgPose$ that correspond to each video frame. 
}
\label{fig:simu}
\end{figure}

%% file: 05_experiments.tex
\section{Experiments}
\label{sec:experiments}
We start by describing the datasets and the metrics we evaluate our method on. We then compare to baseline methods in~\cref{subsec:experiments:comparison}. Next, we ablate the \loopN component in~\cref{subsec:experiments:ablation}. For additional implementation details, we refer the reader to our supplementary material.

\noindent\textbf{Datasets.} 
We carefully choose the evaluation datasets to be those that have significant inter-person interaction, where purely kinematics-based approaches tend to fail. 
To this end, we evaluate our method on three datasets with increasing levels of interaction. \textbf{CHI3D}~\cite{Fieraru_2020_CVPR} contains mild interactions, while more intense interactions in \textbf{Hi4D}~\cite{yin2023hi4d}, and significant interaction and occlusion in \textbf{ExPI}~\cite{guo2021multi}. ExPI also features extreme poses and highly dynamic motion. 
There are other datasets containing multiple interacting people, such as MuPoTS-3D~\cite{singleshot}, ShakeFive2~\cite{shakeFive2}, and MultiHuman~\cite{multiHumanData}. However, they do not contain close interactions in the same amount as the three above.

CHI3D contains 127 motion sequences for each of the 5 pairs of subjects (3 train, 2 test) interacting in everyday actions such as posing (for a photograph), pushing, hugging, etc. CHI3D is captured with cameras from four different views. Each motion sequence is annotated with the action label together with ground truth 3D poses in a world coordinate system in the SMPL format.

The Hi4D dataset is captured with up to 8 cameras at different locations. 
It contains 100 short motion sequences with close interaction and high contact ratio between the subjects performing diverse actions, such as hugging, posing, dancing, and playing sports. It includes 20 unique pairs of participants.

ExPI is the most challenging dataset that contains subjects performing 16 complicated two-person dance routines and, thus, presents highly dynamic sequences with a high amount of contact. Each of the aerials is repeated five times and in total, it contains 115 motion sequences. The data is collected with 60+ synchronized cameras and a motion capture system.

\noindent\textbf{Evaluation metrics.} 
We report a variety of metrics in two categories: (1) pose metrics, which measure both the pose accuracy and smoothness, and (2) physics metrics, which measure the physics plausibility of the motion sequence. 
For the first category, we follow SLAHMR~\cite{ye2023slahmr} and report the World PA First - MPJPE (\textit{W-MPJPE}), which reports the MPJPE~\cite{h36m_pami} after aligning the \textit{first} coordinate frames of the prediction and the ground truth, as well as \textit{PA-MPJPE (joint)}, which reports the MPJPE error after \textit{jointly} aligning the predictions of all the people in \textit{all} frames with the ground truth poses. We also report the acceleration error (\textit{Acc. Error}), which measures the acceleration difference between the ground truth and the estimated motion. In practice, it serves as a measure for motion jittering. For the second category, we report the foot skating (\textit{Skating}) and ground penetration (\textit{Gnd Pen.}) metrics following~\cite{humor,Luo2022EmbodiedSH}. We also report the amount of inter-person penetration (\textit{Pen.}). Specifically, we compute the signed distance function (SDF) values for each person and evaluate each penetrating vertex of one person into the other. We then report the cumulative values of this, \ie we report the sum of SDF values for all penetrating vertices averaged \textit{per person} for the entire sequence.
We report the results in meters, except the \textit{Pen.} metric which is in mm. Please see the Supp. Mat. for more details on the evaluation protocol.

\input{figures/curves}

\subsection{Comparison with Baselines}
\label{subsec:experiments:comparison}
To the best of our knowledge, there is no previous approach that uses a physics simulator to estimate 3D motion for multiple people. We hence extend EmbPose~\cite{Luo2022EmbodiedSH} to take 2D keypoints as input and operate in the simulation environment for all people at the same time. Specifically, we extend their model to have $N$ identical but separate branches (one per person). Each branch produces an initial estimate using the kinematics-based policy $\pi_{KIN}$, and an action signal to drive the simulation with the policy $\pi$. We also adapt the simulation to include $N$ humanoid agents with different poses and body shapes that move independently but simultaneously. We refer to the adapted method as EmbPose-MP. Additionally, we compare to SLAHMR~\cite{ye2023slahmr} as the state-of-the-art multi-person motion estimation method, which is purely kinematics-based.

We report the results in~\cref{tab:sota}. On Hi4D and ExPI, our method outperforms SLAHMR on all \physics metrics, while inferior in terms of acceleration. This is because SLAHMR directly optimizes for smooth motions, which results in lower acceleration errors.

On CHI3D, our method outperforms both SLAHMR and EmbPose-MP in terms of pose metrics. These results are encouraging as they show that enforcing physics compliance can not only enhance the physical plausibility of the estimated motion, but can also lead to more accurate poses. Pose improvement happens especially in cases where people have both feet in contact with the ground. In the simulation, the ground is taken as a hard constraint and thus when the agent moves, it cannot penetrate it, resulting in more realistic poses. This effect has also been reported before~\cite{yuan2021simpoe,shimadaPhysCap}.
The physics simulation also helps improve the poses in cases where the two people are in close proximity, allowing for better spatial placement as body meshes cannot penetrate. This is reflected mostly in the W-MPJPE metric, and happens when people are, e.g., touching or hugging, as shown in~\cref{fig:results_hi4d} and~\cref{fig:results_expi}. Thus, our method improves the motion estimation in these cases by correcting the inter-person penetrations, which often exist in SLAHMR poses.

\input{tables/sota}
\input{figures/results_1}
\input{figures/results_2}

\noindent \textbf{Results on CHI3D.}
For CHI3D, we see that our system outperforms SLAHMR both in penetration (Pen. and Gnd Pen.) and pose metrics. However, for this dataset, SLAHMR has a better skating score, which we hypothesize is due to the fact that CHI3D contains less dynamic motions than the others, and most of the time both people keep their feet on the ground (as opposed to, \eg in ExPI). This results in better ground plane estimation making it easier for SLAHMR to deal with skating. 

\noindent \textbf{Results on Hi4D and ExPI.}
For Hi4D, we see that EmbPose-MP does better on penetration metrics than our model, while the pose metrics are worse. This is due to poor estimation of both global spatial placement and individual poses. Because EmbPose-MP cannot handle inter-person occlusion, poses where the people are close together are not well captured by it and tend to be estimated farther away from each other, when they should in fact be closer together or in contact. As a consequence, penetrations are naturally less likely to occur.
 
In contrast, our model is able to capture people in close proximity while not breaking the laws of physics (see~\cref{fig:results_hi4d} and~\cref{fig:results_expi}). This is reflected in better pose while still achieving good penetration metrics.
Note that penetration reduces drastically in comparison to the purely kinematic baseline which recovers poses accurately but presents high penetration. Moreover, skating and ground penetration are corrected (see Gnd. Pen. and Skating metrics) \wrt the kinematic method. The baseline has better skating scores than our model due to our \loopN which slightly introduces skating as explained in ~\cref{subsec:experiments:ablation}. 
For both datasets, we observe that ground penetration is worse for the baseline as it presents erratic estimated poses and also jittery motion.
For these datasets, we observe pose metric values close to the ones obtained with SLAHMR but with slight differences. This is caused mainly by the type of motion present in each dataset. Hi4D and ExPI, in contrast to CHI3D, contain more dynamic motion, which in some cases can be harder to match for the simulated agent.

\subsection{Ablation Study}
\label{subsec:experiments:ablation}
We perform an ablation to study the effect of the \loopN component in our system, whose results are reported in~\cref{fig:curves}.
We ablate on different values of $\nloop$ and report the performance of: 
(i) the basic version of our approach (Loop1) where we use kinematic estimates plus the physics simulator and 
(ii) our method with \loopN variant for different values of $\nloop$, where $\nloop > 1$. 
The measured metrics are: Gnd Pen., Skating, Pen., W-MPJPE, PA-MPJPE, and Acc. Error. In~\cref{fig:curves}, we show plots of the metrics in two groups: physics in~\cref{fig:curves_a} and pose in~\cref{fig:curves_b} to better analyze the effects and trends on these when $\nloop$ is changed. 
We choose to perform our ablation study on Hi4D as it is the most representative among the datasets. It includes both mild and dynamic motion and at the same time poses where people are very close spatially. 

Our \loopN component, which composes the full system, helps the simulated poses to better match the kinematic reference poses especially for highly dynamic motions such as the ones present in ExPI, see ~\cref{sec:meth:simu}.
We see that with a correctly chosen value of $\nloop$, the policy is able to better match the reference poses. 

The \loopN component gets the simulated poses closer to the kinematic reference pose, however, the best value for $\nloop$ to ensure a better match, depends on the particular motion. Nevertheless, we observe that $\nloop=2$ works well for all the datasets, striking a good balance between pose and \physics metrics. 
In the curves in ~\cref{fig:curves_a}, we see that Acc Error., Gnd Pen. and Pen. almost stay constant with small variations when $\nloop$ changes. This is different for Skating as it increases with $\nloop$. This is due to the fact that as $\nloop$ increases, the pose change between frames is potentially larger, thus for a given value of Gnd Pen., increases Skating. In ~\cref{fig:curves_b}, we see that for $\nloop=2$ the pose errors decrease, however, for values where $\nloop>2$ the error starts to increase. The results shown in ~\cref{tab:sota} are calculated using $\nloop=2$.

\input{figures/more_people}

\subsection{Scaling to More People}
\label{subsec:experiments:scaling}
There is no inherent limitation preventing our model from scaling to scenes with more people. However, most datasets that contain close interactions only consider two people since that is the unit of such interactions. Current datasets with more people do not capture close interactions.
To showcase our method's capability, we apply it to videos with more than two people (see \cref{fig:more_people}). Note that our model reliably captures the human's pose and spatial placement also in these scenarios.

%% file: figures/curves.tex
\begin{figure*}
    \centering
    \begin{subfigure}[b]{0.5\textwidth}
        \includegraphics[width=\textwidth]{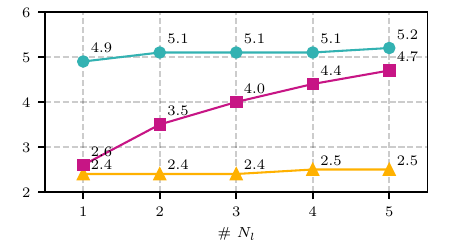}
    \end{subfigure}%
    \begin{subfigure}[b]{0.5\textwidth}
        \includegraphics[width=\textwidth]{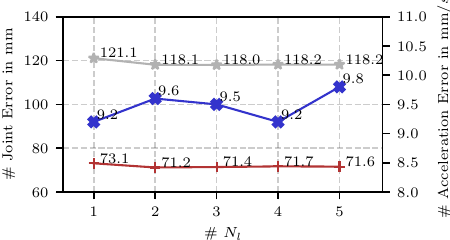}
    \end{subfigure}\\[0.3em]
    \begin{subfigure}[t]{0.5\textwidth}
        \includegraphics[width=\columnwidth]{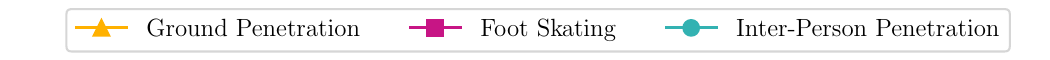}
        \caption{Effect on Physics-based Metrics}
        \label{fig:curves_a}
    \end{subfigure}%
    \begin{subfigure}[t]{0.5\textwidth}
        \includegraphics[width=\columnwidth]{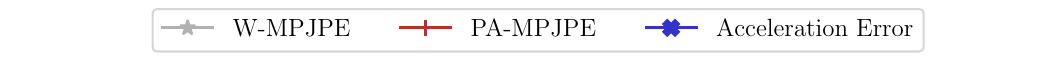}
        \caption{Effect on Pose Metrics}
        \label{fig:curves_b}
    \end{subfigure}
\caption{
\textbf{Effect of \loopN component for different values of $\nloop$.} We study the effect of different values of $\nloop=\{1, ..., 5\}$ on both (a) physics and (b) pose metrics. We report Inter-Person Penetration (measured in m.), the Ground Penetration (measured in mm), the Floor Skating (measured in mm), the W-MPJPE and PA-MPJPE (measured in mm) and the Acceleration Error (measured in mm/s\textsuperscript{2}). We choose $\nloop=2$ for the rest of the experiments as it provides a good balance between physics and pose metrics, see ~\cref{subsec:experiments:ablation}. Note that we scale Pen. metric by a factor of 1/10 to fit the graph. To see the table for these numbers refer to the Supp. Mat.}
\vspace*{-0.2cm}
\label{fig:curves}
\end{figure*}

%% file: tables/sota.tex
\begin{table*}[]
\centering
\resizebox{0.95\linewidth}{!}{%
\begin{tabular}{@{}cl cccc Hcc@{}}
\toprule
\multicolumn{1}{l}{}   & Method          &Pen.\doA  & Gnd Pen.\doA  & Skating\doA  & Acc. Error\doA &  WA-MPJPE\doA  & W-MPJPE\doA  & PA-MPJPE (joint)\doA   \\ \midrule
\multirow{3}{*}{CHI3D} & SLAHMR~\cite{ye2023slahmr}           & 139.3         & 4.4          & \textbf{1.0} & \textbf{6.5} & {\ul 100.9}    & {\ul 177.1}    & {\ul 83.5}     \\
                       & EmbPose-MP~\cite{Luo2022EmbodiedSH}         & {\ul 40.2}    & \textbf{2.6} & 2.8          & 7.7          & 126.7          & 214.7          & 96.5           \\
                       & Ours            & \textbf{18.7} & {\ul 3.2}    & {\ul 2.7}    & {\ul 7.4}    & \textbf{98.1}  & \textbf{174.7} & \textbf{80.4}  \\ \midrule
\multirow{3}{*}{Hi4D}  & SLAHMR~\cite{ye2023slahmr}           & 367.3         & 12.2         & 4.9          & \textbf{6.9} & \textbf{80.9}  & {\ul 121.6}    & \textbf{69.1}  \\
                       & EmbPose-MP~\cite{Luo2022EmbodiedSH}  & \textbf{39.8} & {\ul 3.8}    & \textbf{1.3} & 12.7         & 115.3          & 148.8          & 92.9           \\
                       & Ours            & {\ul 51.1}    & \textbf{2.4} & {\ul 3.5}    & {\ul 9.6}    & {\ul 84.8}     & \textbf{118.1} & {\ul 71.2}     \\ \midrule
\multirow{3}{*}{ExPI}  & SLAHMR~\cite{ye2023slahmr}           & 567.3         & 18.6         & 5.4          & \textbf{8.2} & \textbf{187.3} & {\ul 263.3}    & \textbf{159.1} \\
                       & EmbPose-MP~\cite{Luo2022EmbodiedSH}         & {\ul 92.1}    & {\ul 0.9}    & \textbf{1.9} & 27.7         & 288.6          & 386.4          & 207.6          \\
                       & Ours            & \textbf{73.0} & \textbf{0.6} & {\ul 2.3}    & {\ul 17.1}   & {\ul 196.3}    & \textbf{250.9} & {\ul 164.3}    \\ \bottomrule
\end{tabular}
} \\[-0.2cm]
\caption{\textbf{Comparison with the state of the art.} We report various metrics on CHI3D~\cite{Fieraru_2020_CVPR}, Hi4D~\cite{yin2023hi4d}, and ExPI~\cite{guo2021multi} datasets. Pose metrics are W-MPJPE and PA-MPJPE (joint) in mm. See ~\cref{subsec:experiments:comparison}.
}
\label{tab:sota}
\end{table*}

%% file: figures/results_1.tex
\begin{figure*}[h!]
\centering
\includegraphics[width=1\linewidth, , trim={0cm 0.0cm 0cm 0}]{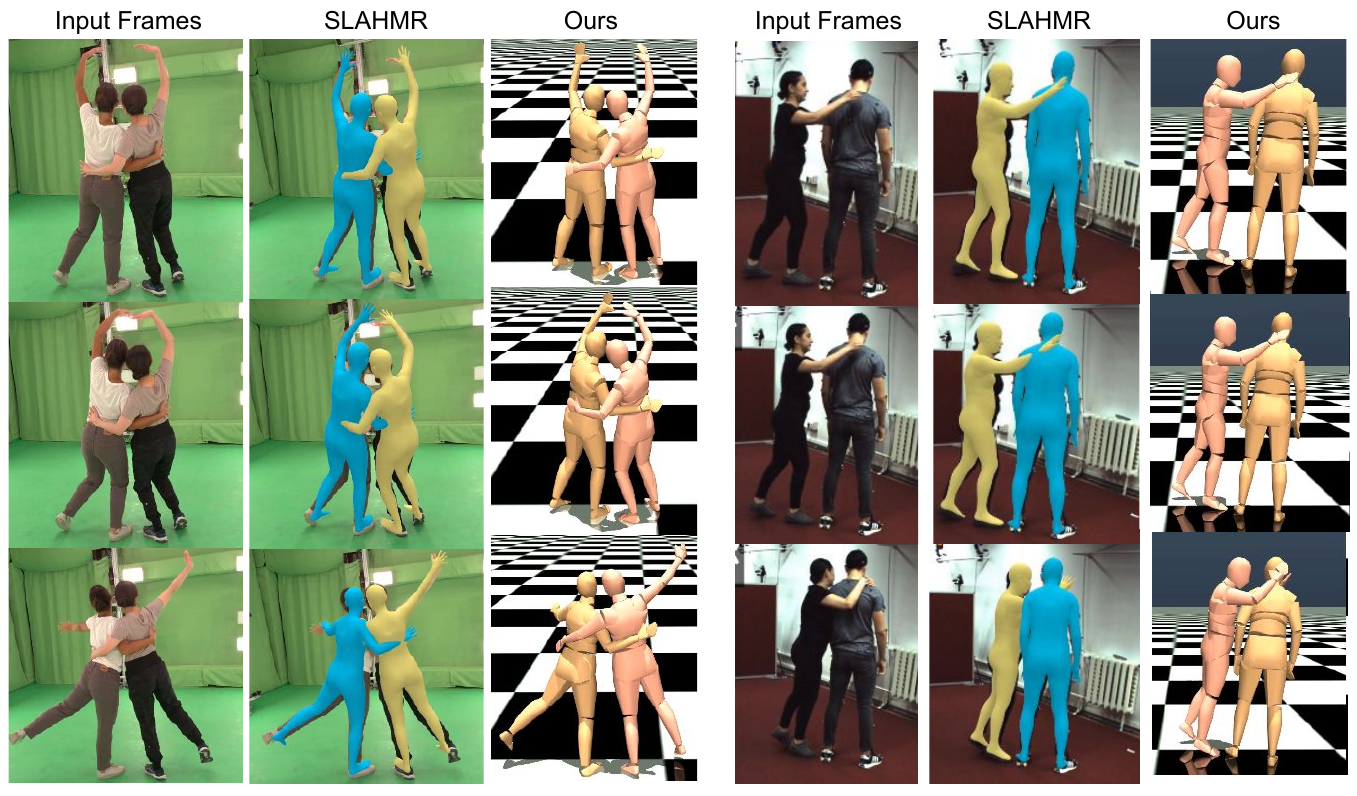} \\[-0.2cm]
\caption{
\textbf{Qualitative results of the proposed approach.} The first three columns (from left to right) are from Hi4D~\cite{yin2023hi4d} and the other three are from CHI3D~\cite{Fieraru_2020_CVPR}. Each row corresponds to one frame of the same sequence. The columns compare the resulting poses at each frame using SLAHMR~\cite{ye2023slahmr} and our method. In these cases of close inter-person interaction, the estimated motion from SLAHMR often has severe inter-person penetrations, while our method is able to eliminate these penetrations through physics-aware correction. 
\vspace*{-0.1cm}
}
\label{fig:results_hi4d}
\end{figure*}

%% file: figures/results_2.tex
\begin{figure*}[h!]
\centering
\includegraphics[width=1\linewidth, trim={0cm 0.0cm 0cm 0}]{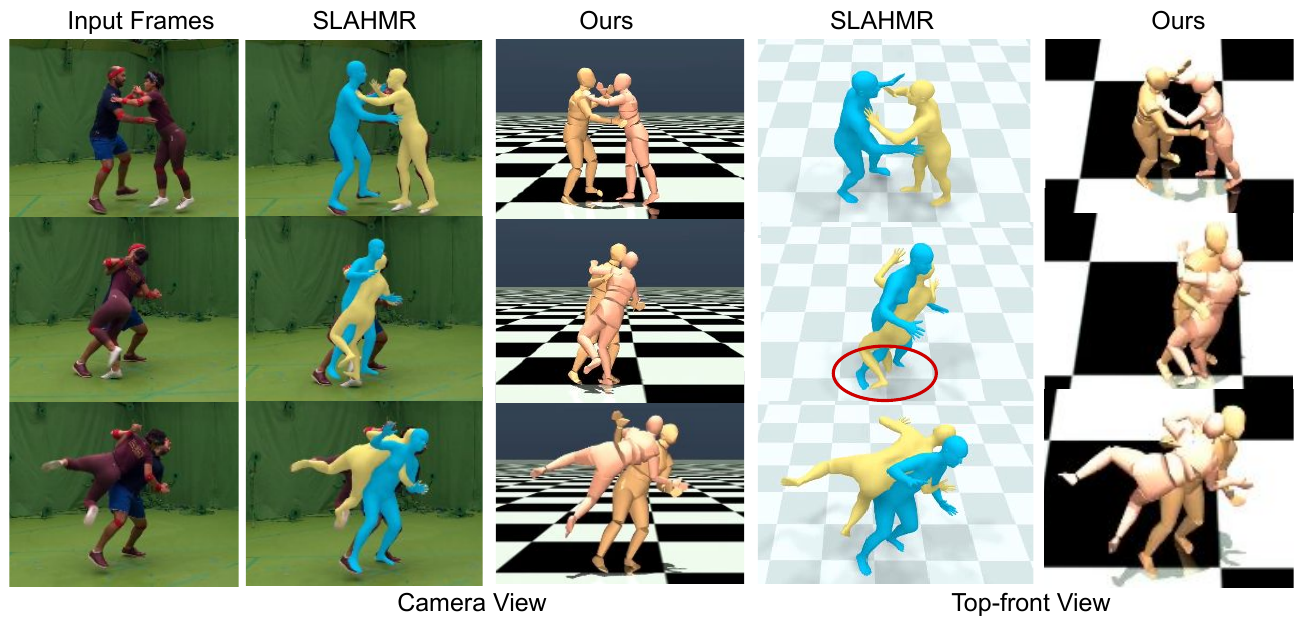} \\[-0.2cm]
\vspace*{0.0cm}
\caption{
\textbf{Effect of the physical constraints on spatial placement.} One key advantage of our method is that complying with physical constraints (e.g., penetration between bodies) helps to improve the spatial placement of the bodies. Here we show results for motion estimated with both the kinematic approach SLAHMR~\cite{ye2023slahmr} and our system. See how the bodies from the kinematic poses overlap and penetrate the ground (red circle in the figure) leading to unrealistic spatial placement. Our method eliminates these penetrations both \wrt the body and the ground. 
\vspace*{-0.2cm}
}
\label{fig:results_expi}
\end{figure*}

%% file: figures/more_people.tex
\begin{figure}[t]
  \centering
  \includegraphics[width=1.0\linewidth]{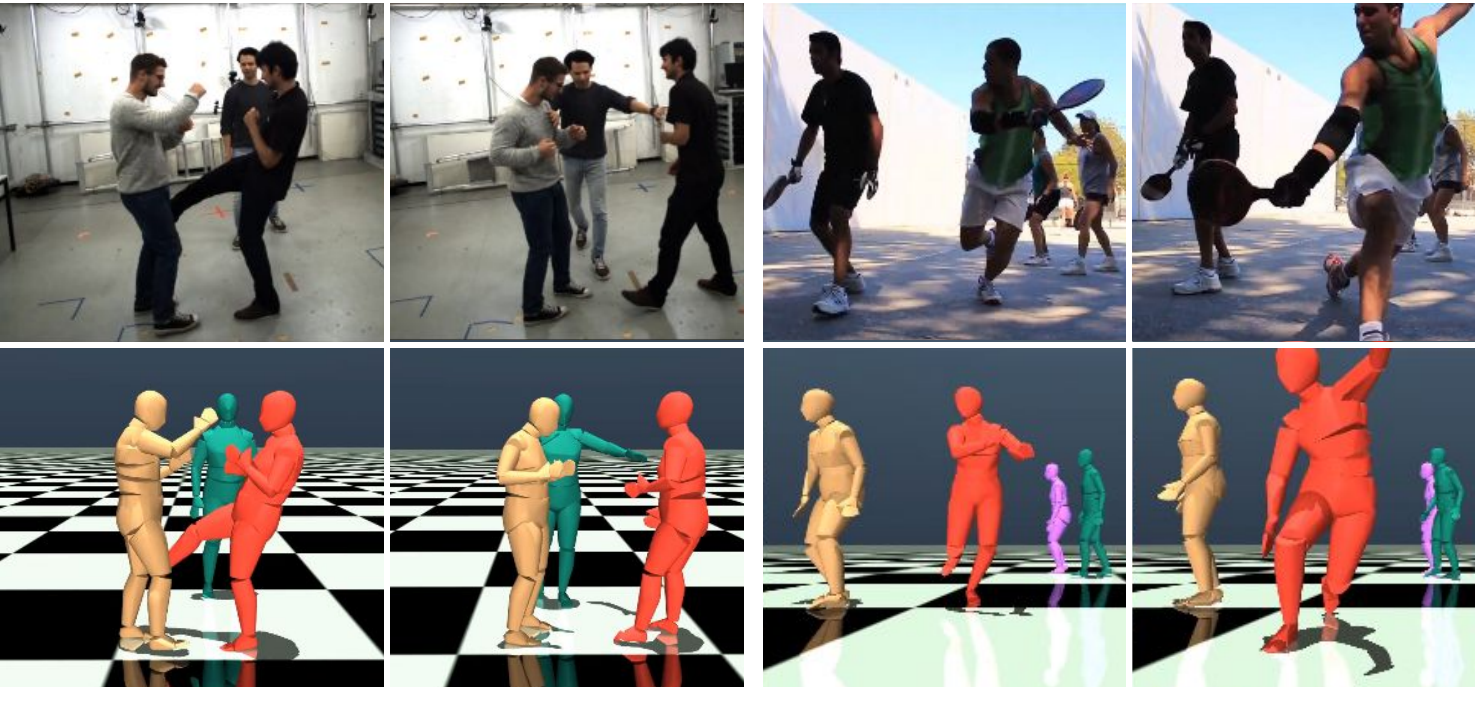}
     \vspace{-5mm}
   \caption{\textbf{Additional results} on videos with three (left) and four (right) people.}
   \label{fig:more_people}
   \vspace{-4mm}
\end{figure}

%% file: 06_conclusion.tex
\section{Discussion}
\label{sec:conclusion}

We propose a method for recovering physically plausible 3D human motion from a monocular RGB video, and in particular for two interacting people. Our approach leverages a fully-featured physics simulator to add constraints to the motion estimation process and to force it to follow the laws of physics.
This allows us to improve the realism of the estimated motion by both avoiding penetration between human bodies and with the ground plane, while also improving the pose in terms of spatial placement. This is corroborated by our experiments on three challenging datasets.

While our method unlocks many new possibilities for generating more realistic and physically plausible motions, some problems remain to be addressed. When 2D keypoint detectors fail, the kinematic-based motion estimation on which we rely as initialization also fails, leading to a decline of pose accuracy. Developing more powerful and robust 2D keypoint detector is hence important in further improving the performance of our method. Please see supplement for failure cases. Another interesting direction is to develop a two-person interaction prior to guide the physics-aware correction process.

%% file: camera_ready.bbl
\begin{thebibliography}{10}\itemsep=-1pt

\bibitem{choi2021beyond}
Hongsuk Choi, Gyeongsik Moon, Ju~Yong Chang, and Kyoung~Mu Lee.
\newblock Beyond static features for temporally consistent 3d human pose and shape from a video.
\newblock In {\em Proceedings of the IEEE/CVF Conference on Computer Vision and Pattern Recognition (CVPR)}, pages 1964--1973, 2021.

\bibitem{Fieraru_2020_CVPR}
Mihai Fieraru, Mihai Zanfir, Elisabeta Oneata, Alin-Ionut Popa, Vlad Olaru, and Cristian Sminchisescu.
\newblock Three-dimensional reconstruction of human interactions.
\newblock In {\em Proceedings of the IEEE/CVF Conference on Computer Vision and Pattern Recognition (CVPR)}, June 2020.

\bibitem{REMIPS}
Mihai Fieraru, Mihai Zanfir, Teodor Szente, Eduard Bazavan, Vlad Olaru, and Cristian Sminchisescu.
\newblock Remips: Physically consistent 3d reconstruction of multiple interacting people under weak supervision.
\newblock In {\em Advances in Neural Information Processing Systems}, 2021.

\bibitem{HMR22023}
Shubham Goel, Georgios Pavlakos, Jathushan Rajasegaran, Angjoo Kanazawa, and Jitendra Malik.
\newblock Humans in 4d: Reconstructing and tracking humans with transformers.
\newblock {\em Proceedings of the IEEE/CVF International Conference on Computer Vision (ICCV)}, 2023.

\bibitem{gartner2022diffphy}
Erik Gärtner, Mykhaylo Andriluka, Erwin Coumans, and Cristian Sminchisescu.
\newblock Differentiable dynamics for articulated 3d human motion reconstruction.
\newblock In {\em Proceedings of the IEEE/CVF Conference on Computer Vision and Pattern Recognition (CVPR)}, 2022.

\bibitem{gartner2022trajectory}
Erik Gärtner, Mykhaylo Andriluka, Hongyi Xu, and Cristian Sminchisescu.
\newblock Trajectory optimization for physics-based reconstruction of 3d human pose from monocular video.
\newblock In {\em Proceedings of the IEEE/CVF Conference on Computer Vision and Pattern Recognition (CVPR)}, 2022.

\bibitem{h36m_pami}
Catalin Ionescu, Dragos Papava, Vlad Olaru, and Cristia Sminchisescu.
\newblock Human3.6m: Large scale datasets and predictive methods for 3d human sensing in natural environments.
\newblock In {\em IEEE. trans. PAMI}, 2014.

\bibitem{jiang2022avatarposer}
Jiaxi Jiang, Paul Streli, Huajian Qiu, Andreas Fender, Larissa Laich, Patrick Snape, and Christian Holz.
\newblock Avatarposer: Articulated full-body pose tracking from sparse motion sensing.
\newblock In {\em European conference on computer vision}, pages 443--460. Springer, 2022.

\bibitem{jiang2020_multiperson}
Wen Jiang, Nikos Kolotouros, Georgios Pavlakos, Xiaowei Zhou, and Kostas Daniilidis.
\newblock Coherent reconstruction of multiple humans from a single image.
\newblock In {\em Proceedings of the IEEE/CVF Conference on Computer Vision and Pattern Recognition (CVPR)}, 2020.

\bibitem{kanazawa_hmr}
Angjoo Kanazawa, Michael~J. Black, David~W. Jacobs, and Jitendra Malik.
\newblock End-to-end recovery of human shape and pose.
\newblock In {\em Proceedings of the IEEE Conference on Computer Vision and Pattern Recognition (CVPR)}, 2018.

\bibitem{hmmr}
Angjoo Kanazawa, Jason~Y Zhang, Panna Felsen, and Jitendra Malik.
\newblock Learning 3d human dynamics from video.
\newblock In {\em Proceedings of the IEEE/CVF Conference on Computer Vision and Pattern Recognition (CVPR)}, pages 5614--5623, 2019.

\bibitem{Khirodkar_2022_CVPR}
Rawal Khirodkar, Shashank Tripathi, and Kris Kitani.
\newblock Occluded human mesh recovery.
\newblock In {\em Proceedings of the IEEE/CVF Conference on Computer Vision and Pattern Recognition (CVPR)}, pages 1715--1725, June 2022.

\bibitem{VIBECVPR2020}
Muhammed Kocabas, Nikos Athanasiou, and Michael~J Black.
\newblock Vibe: Video inference for human body pose and shape estimation.
\newblock In {\em Proceedings of the IEEE Conference on Computer Vision and Pattern Recognition (CVPR)}, 2020.

\bibitem{kocabas2024pace}
Muhammed Kocabas, Ye Yuan, Pavlo Molchanov, Yunrong Guo, Michael~J. Black, Otmar Hilliges, Jan Kautz, and Umar Iqbal.
\newblock Pace: Human and motion estimation from in-the-wild videos.
\newblock In {\em 3DV}, 2024.

\bibitem{lee2023questenvsim}
Sunmin Lee, Sebastian Starke, Yuting Ye, Jungdam Won, and Alexander Winkler.
\newblock Questenvsim: Environment-aware simulated motion tracking from sparse sensors.
\newblock In {\em ACM SIGGRAPH 2023 Conference Proceedings}, pages 1--9, 2023.

\bibitem{li2022dnd}
Jiefeng Li, Siyuan Bian, Chao Xu, Gang Liu, Gang Yu, and Cewu Lu.
\newblock D\&d: Learning human dynamics from dynamic camera.
\newblock In {\em European Conference on Computer Vision (ECCV)}, 2022.

\bibitem{luo2020_motion}
Zhengyi Luo, S~Alireza Golestaneh, and Kris~M Kitani.
\newblock 3d human motion estimation via motion compression and refinement.
\newblock In {\em Proceedings of the Asian Conference on Computer Vision}, 2020.

\bibitem{Luo2021kinpoly}
Zhengyi Luo, Ryo Hachiuma, Ye Yuan, and Kris Kitani.
\newblock Dynamics-regulated kinematic policy for egocentric pose estimation.
\newblock In {\em Advances in Neural Information Processing Systems}, 2021.

\bibitem{Luo2022EmbodiedSH}
Zhengyi Luo, Shun Iwase, Ye Yuan, and Kris Kitani.
\newblock Embodied scene-aware human pose estimation.
\newblock In {\em Advances in Neural Information Processing Systems}, 2022.

\bibitem{AMASS:ICCV:2019}
Naureen Mahmood, Nima Ghorbani, Nikolaus~F. Troje, Gerard Pons-Moll, and Michael~J. Black.
\newblock Amass: Archive of motion capture as surface shapes.
\newblock In {\em Proceedings of the IEEE/CVF International Conference on Computer Vision (ICCV)}, 2019.

\bibitem{singleshot}
Dushyant Mehta, Oleksandr Sotnychenko, Franziska Mueller, Weipeng Xu, Srinath Sridhar, Gerard Pons-Moll, and Christian Theobalt.
\newblock Single-shot multi-person 3d pose estimation from monocular rgb.
\newblock In {\em International Conference on 3D Vision (3DV)}, pages 120--130. IEEE, 2018.

\bibitem{muller2023buddi}
Lea M{\"u}ller, Vickie Ye, Georgios Pavlakos, Michael Black, and Angjoo Kanazawa.
\newblock Generative proxemics: A prior for 3d social interaction from images.
\newblock {\em arXiv preprint arXiv:2306.09337}, 2023.

\bibitem{smplx}
Georgios Pavlakos, Vasileios Choutas, Nima Ghorbani, Timo Bolkart, Ahmed~A. Osman, Dimitrios Tzionas, and Michael~J. Black.
\newblock Expressive body capture: 3d hands, face, and body from a single image.
\newblock In {\em Proceedings of the IEEE/CVF Conference on Computer Vision and Pattern Recognition (CVPR)}, 2019.

\bibitem{pavlakos2022multishot}
Georgios Pavlakos, Jitendra Malik, and Angjoo Kanazawa.
\newblock Human mesh recovery from multiple shots.
\newblock In {\em Proceedings of the IEEE/CVF Conference on Computer Vision and Pattern Recognition (CVPR)}, 2022.

\bibitem{2018-TOG-SFV}
Xue~Bin Peng, Angjoo Kanazawa, Jitendra Malik, Pieter Abbeel, and Sergey Levine.
\newblock Sfv: Reinforcement learning of physical skills from videos.
\newblock {\em ACM Trans. Graph.}, 37(6), Nov. 2018.

\bibitem{humor}
Davis Rempe, Tolga Birdal, Aaron Hertzmann, Jimei Yang, Srinath Sridhar, and Leonidas~J. Guibas.
\newblock Humor: 3d human motion model for robust pose estimation.
\newblock Proceedings of the IEEE/CVF International Conference on Computer Vision (ICCV), 2021.

\bibitem{smplh2022}
Javier Romero, Dimitrios Tzionas, and Michael~J Black.
\newblock Embodied hands: Modeling and capturing hands and bodies together.
\newblock {\em ACM Transactions on Graphics, (Proc. SIGGRAPH Asia)}, 2022.

\bibitem{schulman2017proximal}
John Schulman, Filip Wolski, Prafulla Dhariwal, Alec Radford, and Oleg Klimov.
\newblock Proximal policy optimization algorithms.
\newblock {\em arXiv preprint arXiv:1707.06347}, 2017.

\bibitem{neuralPhysCap}
Soshi Shimada, Vladislav Golyanik, Weipeng Xu, Patrick P\'{e}rez, and Christian Theobalt.
\newblock Neural monocular 3d human motion capture with physical awareness.
\newblock {\em ACM Transactions on Graphics}, 40(4), aug 2021.

\bibitem{shimadaPhysCap}
Soshi Shimada, Vladislav Golyanik, Weipeng Xu, and Christian Theobalt.
\newblock Physcap: Physically plausible monocular 3d motion capture in real time.
\newblock {\em ACM Transactions on Graphics}, 39(6), dec 2020.

\bibitem{ROMP}
Yu Sun, Qian Bao, Wu Liu, Yili Fu, Black Michael~J., and Tao Mei.
\newblock Monocular, one-stage, regression of multiple 3d people.
\newblock In {\em Proceedings of the IEEE/CVF International Conference on Computer Vision (ICCV)}, October 2021.

\bibitem{TRACE}
Yu Sun, Qian Bao, Wu Liu, Tao Mei, and Michael~J. Black.
\newblock {TRACE: 5D Temporal Regression of Avatars with Dynamic Cameras in 3D Environments}.
\newblock In {\em Proceedings of the IEEE/CVF Conference on Computer Vision and Pattern Recognition (CVPR)}, 2023.

\bibitem{BEV}
Yu Sun, Wu Liu, Qian Bao, Yili Fu, Tao Mei, and Michael~J. Black.
\newblock Putting people in their place: Monocular regression of {3D} people in depth.
\newblock In {\em IEEE/CVF Conf.~on Computer Vision and Pattern Recognition (CVPR)}, June 2022.

\bibitem{todorov2012mujoco}
Emanuel Todorov, Tom Erez, and Yuval Tassa.
\newblock Mujoco: A physics engine for model-based control.
\newblock In {\em 2012 IEEE/RSJ International Conference on Intelligent Robots and Systems}, pages 5026--5033. IEEE, 2012.

\bibitem{ugrinovic2021depthsize}
Nicolas Ugrinovic, Adria Ruiz, Antonio Agudo, Alberto Sanfeliu, and Francesc Moreno-Noguer.
\newblock Body size and depth disambiguation in multi-person reconstruction from single images.
\newblock 2021.

\bibitem{shakeFive2}
Coert van Gemeren, Ronald Poppe, and Remco~C. Veltkamp.
\newblock Spatio-temporal detection of fine-grained dyadic human interactions.
\newblock In Mohamed Chetouani, Jeffrey Cohn, and Albert~Ali Salah, editors, {\em Human Behavior Understanding}, 2016.

\bibitem{guo2021multi}
Guo Wen, Bie Xiaoyu, Alameda-Pineda Xavier, and Moreno-Noguer Francesc.
\newblock Multi-person extreme motion prediction.
\newblock In {\em Proceedings of the IEEE/CVF Conference on Computer Vision and Pattern Recognition (CVPR)}, 2022.

\bibitem{winkler2022questsim}
Alexander Winkler, Jungdam Won, and Yuting Ye.
\newblock Questsim: Human motion tracking from sparse sensors with simulated avatars.
\newblock In {\em SIGGRAPH Asia 2022 Conference Papers}, pages 1--8, 2022.

\bibitem{ye2023slahmr}
Vickie Ye, Georgios Pavlakos, Jitendra Malik, and Angjoo Kanazawa.
\newblock Decoupling human and camera motion from videos in the wild.
\newblock In {\em Proceedings of the IEEE/CVF Conference on Computer Vision and Pattern Recognition (CVPR)}, June 2023.

\bibitem{yin2023hi4d}
Yifei Yin, Chen Guo, Manuel Kaufmann, Juan Zarate, Jie Song, and Otmar Hilliges.
\newblock Hi4d: 4d instance segmentation of close human interaction.
\newblock In {\em Proceedings of the IEEE/CVF Conference on Computer Vision and Pattern Recognition (CVPR)}, 2023.

\bibitem{MovingCam2021}
Ri Yu, Hwangpil Park, and Jehee Lee.
\newblock Human dynamics from monocular video with dynamic camera movements.
\newblock {\em ACM Trans. Graph.}, 40(6), 2021.

\bibitem{yuan2022glamr}
Ye Yuan, Umar Iqbal, Pavlo Molchanov, Kris Kitani, and Jan Kautz.
\newblock Glamr: Global occlusion-aware human mesh recovery with dynamic cameras.
\newblock In {\em Proceedings of the IEEE/CVF Conference on Computer Vision and Pattern Recognition (CVPR)}, 2022.

\bibitem{yuan2020residual}
Ye Yuan and Kris Kitani.
\newblock Residual force control for agile human behavior imitation and extended motion synthesis.
\newblock In {\em Advances in Neural Information Processing Systems}, 2020.

\bibitem{yuan2021simpoe}
Ye Yuan, Shih-En Wei, Tomas Simon, Kris Kitani, and Jason Saragih.
\newblock Simpoe: Simulated character control for 3d human pose estimation.
\newblock In {\em Proceedings of the IEEE/CVF Conference on Computer Vision and Pattern Recognition (CVPR)}, 2021.

\bibitem{multiHumanData}
Yuxiang Zhang, Zhe Li, Liang An, Mengcheng Li, Tao Yu, and Yebin Liu.
\newblock Light-weight multi-person total capture using sparse multi-view cameras.
\newblock In {\em Proceedings of the IEEE/CVF International Conference on Computer Vision (ICCV)}, 2021.

\bibitem{zheng2023realistic}
Xiaozheng Zheng, Zhuo Su, Chao Wen, Zhou Xue, and Xiaojie Jin.
\newblock Realistic full-body tracking from sparse observations via joint-level modeling.
\newblock In {\em Proceedings of the IEEE/CVF International Conference on Computer Vision}, pages 14678--14688, 2023.

\end{thebibliography}
